\def\year{2018}\relax
\documentclass[letterpaper]{article}
\usepackage{aaai18}
\usepackage{times}
\usepackage{helvet}
\usepackage{courier}
\usepackage{url}
\usepackage{graphicx}
\usepackage{times}
\usepackage{epsfig}
\usepackage{amsmath}
\usepackage{amssymb}
\usepackage{subfigure}
\usepackage{url}
\usepackage{calc}
\begin{document}
\title{End-to-End United Video Dehazing and Detection}
\author{Boyi Li$ ^{1}\thanks{The work was done at Microsoft Research Asia.}$, Xiulian Peng$ ^{2}$, Zhangyang Wang$ ^{3} $, Jizheng Xu$ ^{2} $, Dan Feng$ ^{1} $\\
	$ ^{1} $Wuhan National Laboratory for Optoelectronics, Huazhong University of Science and Technology\\
	$ ^{2} $Microsoft Research, Beijing, China\\
	$ ^{3} $Department of Computer Science and Engineering, Texas A\&M University\\
	{\tt\small boyilics@gmail.com,xipe@microsoft.com,atlaswang@tamu.edu,jzxu@microsoft.com,dfeng@hust.edu.cn}
}
\maketitle
\begin{abstract}
The recent development of CNN-based image dehazing has revealed the effectiveness of end-to-end modeling. However, extending the idea to end-to-end video dehazing has not been explored yet. In this paper, we propose an \textit{End-to-End Video Dehazing Network} (\textbf{EVD-Net}), to exploit the temporal consistency between consecutive video frames. A thorough study has been conducted over a number of structure options, to identify the best temporal fusion strategy. Furthermore, we build an \textit{End-to-End United Video Dehazing and Detection Network} (\textbf{EVDD-Net}), which concatenates and jointly trains EVD-Net with a video object detection model. The resulting augmented end-to-end pipeline has demonstrated much more stable and accurate detection results in hazy video.
\end{abstract}

\section{Introduction}

The removal of haze from visual data captured in the wild has been attracting tremendous research interests, due to its profound application values in outdoor video surveillance, traffic monitoring and autonomous driving, and so on \cite{tan2008visibility}. In principle, the generation of hazy visual scene observations follows a known physical model (to be detailed next), and the estimation of key physical parameters, i.e., the atmospheric light magnitude and transmission matrix, become the core step in solving haze removal as an inverse problem \cite{he2011single,fattal2014dehazing,berman2016non}. Recently, the prosperity of convolutional neural networks (CNNs) \cite{krizhevsky2012imagenet} has led to many efforts paid to CNN-based single image dehazing \cite{ren2016single,cai2016dehazenet,ICCV17a}. Among them, DehazeNet \cite{cai2016dehazenet} and MSCNN \cite{ren2016single} focused on predicting the most important parameter, transmission matrix, from image inputs using CNNs, then generating clean images by the physical model. Lately, AOD-Net \cite{ICCV17a} was the first model to introduce a light-weight end-to-end dehazing convolutional neural network by re-formulating the physical formula. However, there have been only a limited amount of efforts in exploring video dehazing, which is the more realistic scenario, either by traditional statistical approaches or by CNNs. 

\begin{figure}
	\begin{center}
		\includegraphics[width=3.3in,height=1.3in]{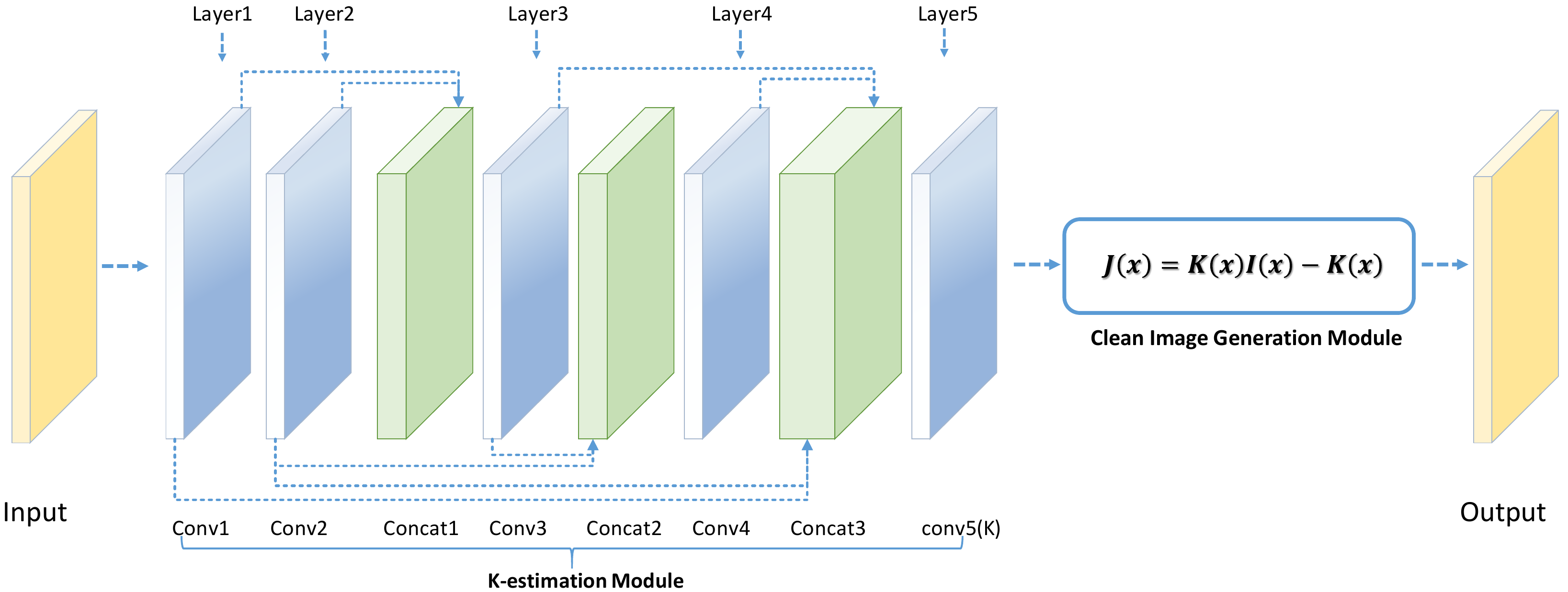}
	\end{center}
	\caption{The AOD-Net architecture for single image dehazing \cite{ICCV17a,li2017all}, which inspires EVD-Net.}
	\vspace{-1em}
	\label{fig:AOD-Net}
\end{figure}

This paper fills in the blank of CNN-based video dehazing by an innovative integration of two important merits in one unified model: (1) we inherit the spirit of training an \textit{end-to-end} model \cite{ICCV17a,wang2016studying}, that directly regresses clean images from hazy inputs without any intermediate step. That is proven to outperform the (sub-optimal) results of multi-stage pipelines; (2) we embrace the video setting by explicitly considering how to embed the temporal coherence between neighboring video frames when restoring the current frame. By an extensive architecture study, we identify the most promising temporal fusion strategy, which is both interpretable from a dehazing viewpoint and well aligned with previous findings \cite{karpathy2014large,kappeler2016video}. We call our proposed model \textit{End-to-End Video Dehazing Network} (\textbf{EVD-Net}). 

Better yet, EVD-Net can be considered as pre-processing for a subsequent high-level computer vision task, and we can therefore jointly train the concatenated pipeline for the optimized high-level task performance in the presence of haze. Using video object detection as a task example, we build the augmented \textit{End-to-End United Video Dehazing and Detection Network} (\textbf{EVDD-Net}), and achieve much more stable and accurate detection results in hazy video.

\section{Related Work}

\begin{figure*}
	\centering
	\subfigure[$I$-level fusion]{
		\includegraphics[width=2.25in,height=1.75in]{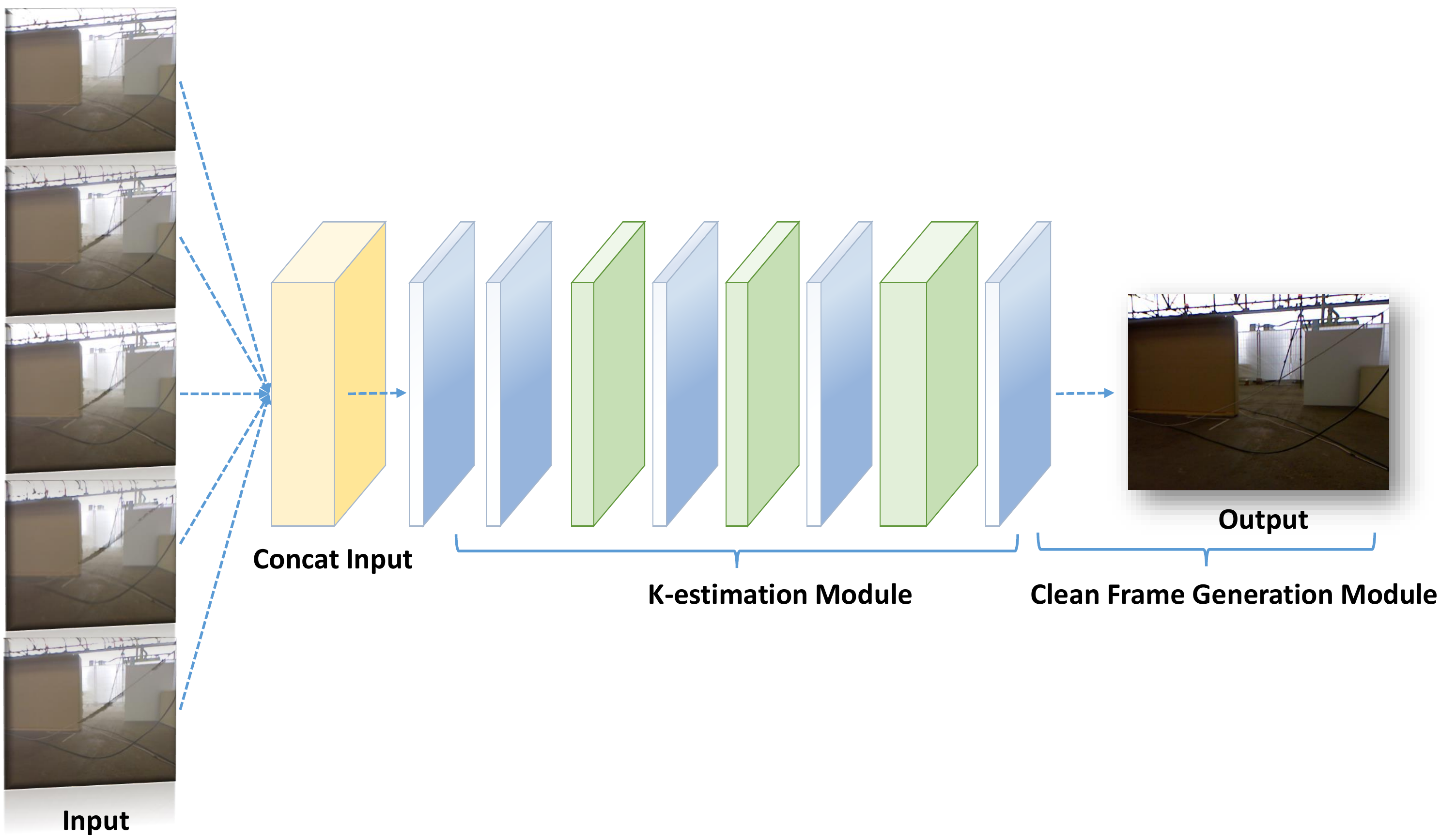}}
	\subfigure[$K$-level fusion]{
		\includegraphics[width=2.25in,height=1.75in]{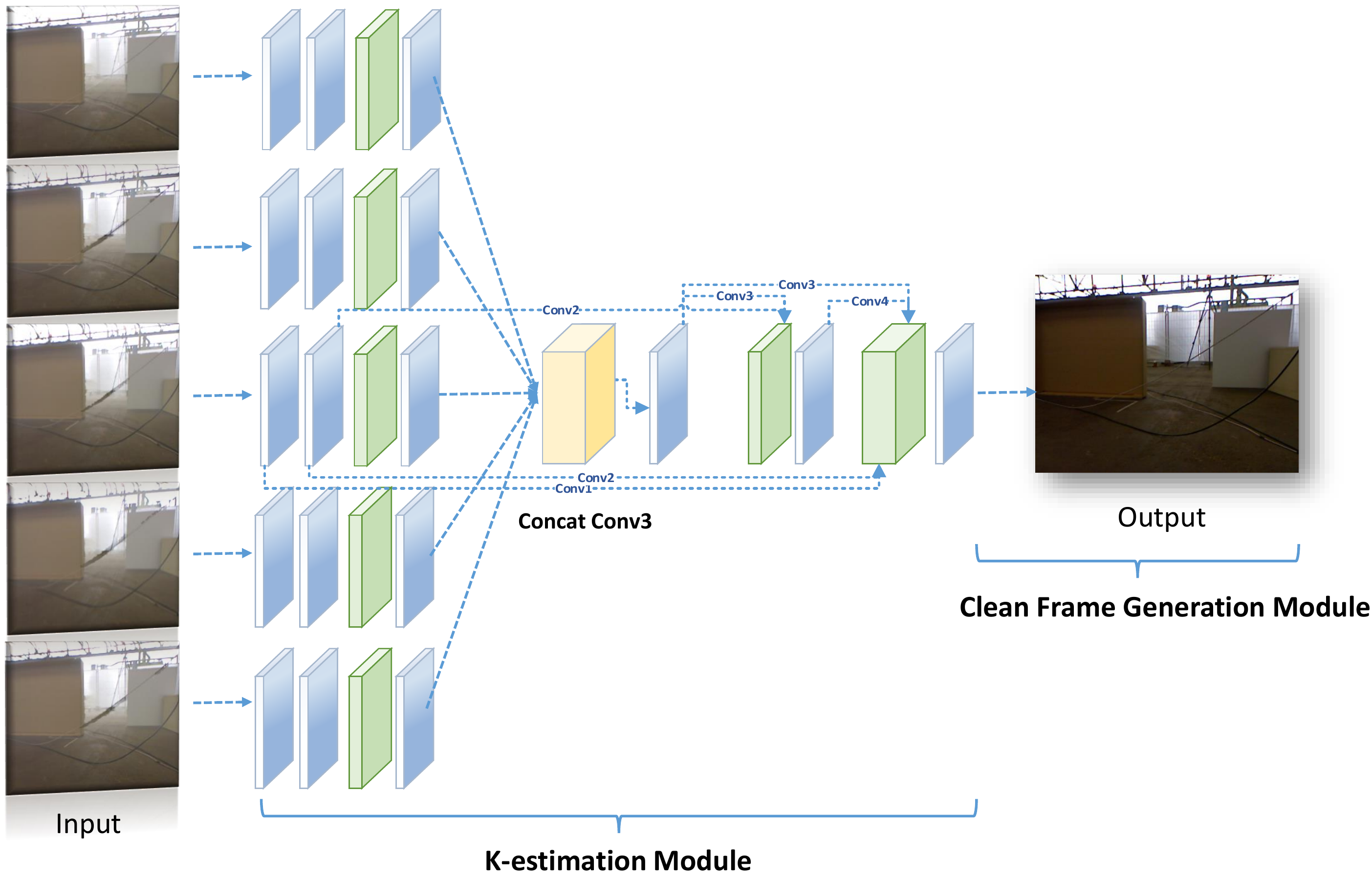}}
	\subfigure[$J$-level fusion]{
		\includegraphics[width=2.25in,height=1.75in]{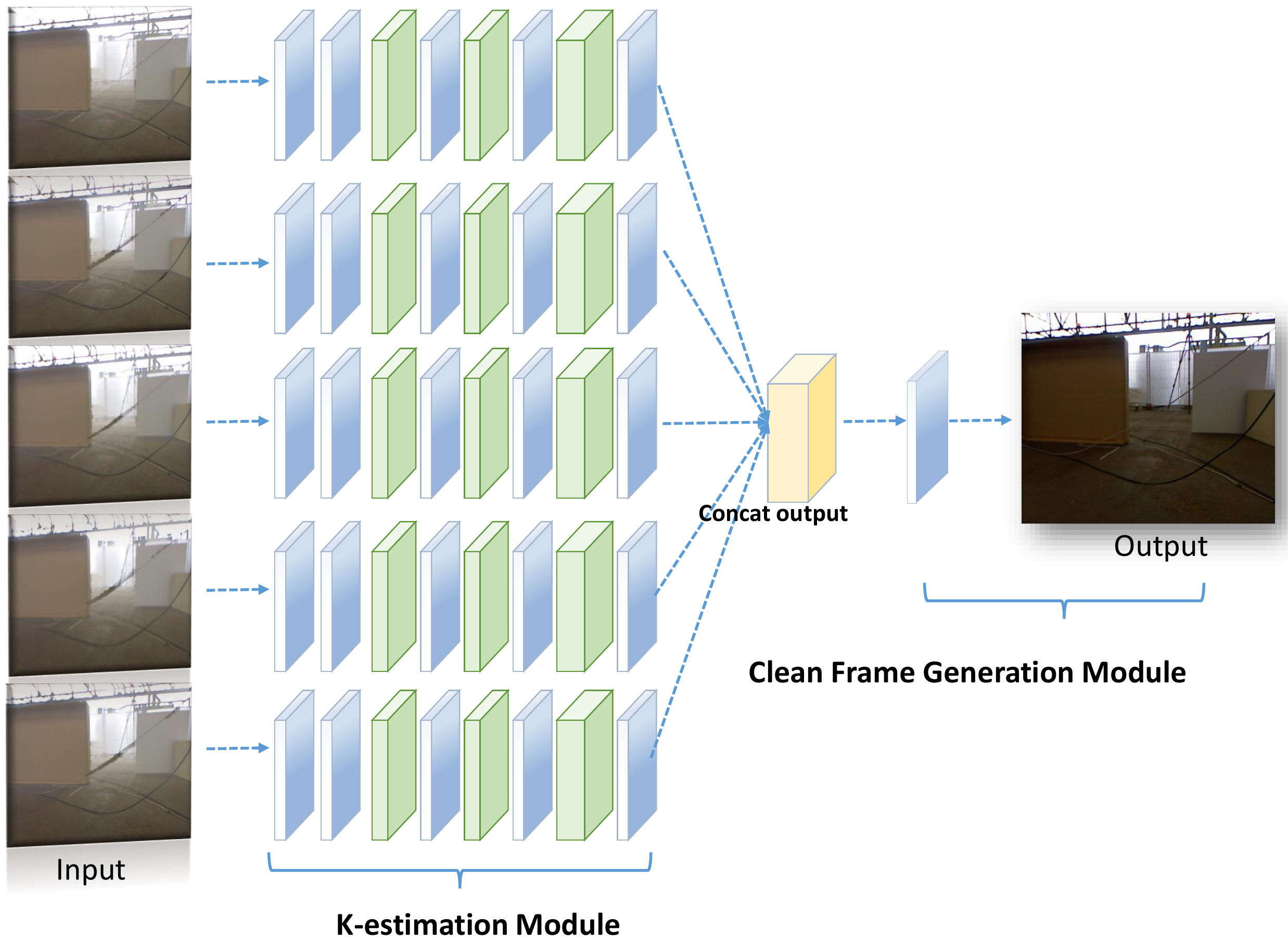}}
	\caption{EVD-Net structure options with 5 consecutive frames as input: (a) $I$-level fusion, where five input frames are concatenated before feeding the first layer; (b) $K$-level fusion, where five input frames are first processed separately in its own column and then concatenated after the some layer during $K$ estimation; (c) $J$-level fusion, where five output images are concatenated.}
	\label{fig:EVD-Net}
\end{figure*}

Previous single image haze removal algorithms focus on the classical \textit{atmospheric scattering model}:
\begin{equation}\label{e1}
I\left( x\right) =J\left( x\right) t\left( x\right) +A\left( 1-t\left( x\right) \right) ,
\end{equation}
where $ I\left( x\right)  $ is observed hazy image, $ J\left( x\right)  $ is the scene radiance (``clean image'') to be recovered. There are two critical parameters: $ A $ denotes the global atmospheric light, and $ t\left( x\right) $ is the transmission matrix defined as:
\begin{equation}\label{e}
t\left( x\right) =e^{-\beta d\left( x\right)},
\end{equation}
where $ \beta $ is the scattering coefficient of the atmosphere, and $ d\left( x\right) $ is the distance between the object and the camera. The clean image can thus be obtained in the inverse way:
\begin{equation}\label{eJ}
J(x) =\frac{1}{t(x)}I(x)-A\frac{1}{t(x)}+A.
\end{equation}

A number of methods \cite{tan2008visibility,fattal2008single,he2011single,meng2013efficient,zhu2015fast} take advantages of natural image statistics as priors, to predict $ A $ and $ t\left( x\right) $ separately from the hazy image $ I\left( x\right) $. Due to the often inaccurate estimation of either (or both), they tend to bring in many artifacts such as non-smoothness, unnatural color tones or contrasts. Many CNN-based methods \cite{cai2016dehazenet,ren2016single} employ CNN as a tool to regress $ t\left( x\right) $ from  $ I\left( x\right) $. With $ A $ estimated using some other empirical methods, they are then able to estimate $J(x)$ by (\ref{eJ}). Notably, \cite{ICCV17a,li2017all} design the \underline{first completely end-to-end} CNN dehazing model
 based on re-formulating (\ref{e1}), which directly generates $J(x)$ from $I(x)$ without any other intermediate step:
 \begin{align}\label{e3}
\begin{split}
J\left( x\right) &=K\left( x\right) I\left( x\right) -K\left( x\right), \text{where} \\
K\left( x\right) &= \dfrac{\frac{1}{t\left( x\right)}(I\left( x\right)-A)+A}{I\left( x\right)-1}.
\end{split}
\end{align}
Both $\frac{1}{t\left( x\right)}$ and $A$ are integrated into the new variable $K\left( x\right)$\footnote{There was a constant bias  $b$ in \cite{ICCV17a,li2017all}, which is omitted here to simplify notations.}.  As shown in Figure \ref{fig:AOD-Net}, the AOD-Net architecture is composed of two modules: a \textit{K-estimation module} consisting of five convolutional layers to estimate $K\left( x\right)$ from $ I\left( x\right)$, followed by a \textit{clean image generation module} to estimate $J\left( x\right)$ from both $K\left( x\right)$ and $ I\left( x\right) $ via (\ref{e3}).  All those above-mentioned methods are designed for single-image dehazing, without taking into account the temporal dynamics in video. 

When it comes to video dehazing, a majority of existing approaches count on post processing to correct temporal inconsistencies, after applying single image dehazing algorithms frame-wise. \cite{kim2013optimized} proposes to inject temporal coherence into the cost function, with a clock filter for speed-up.
\cite{li2015simultaneous} jointly estimates the scene depth and recovers the clear latent images from a foggy video sequence. \cite{chen2016robust} presents an image-guided, depth-edge-aware smoothing algorithm to refine the transmission matrix, and uses Gradient Residual Minimization to recover the haze-free images. \cite{cai2016stmrf} designs a spatio-temporal optimization for real-time video dehazing. But as our experiments will show, those relatively simple and straightforward video dehazing approaches may not be even able to outperform the sophisticated CNN-based single image dehazing models. The observation reminds us that the utility of temporal coherence must be coupled with more advanced model structures (such as CNNs) for the further boost of video dehazing performance.

\begin{figure}
	\includegraphics[width=3.5in,height=3in]{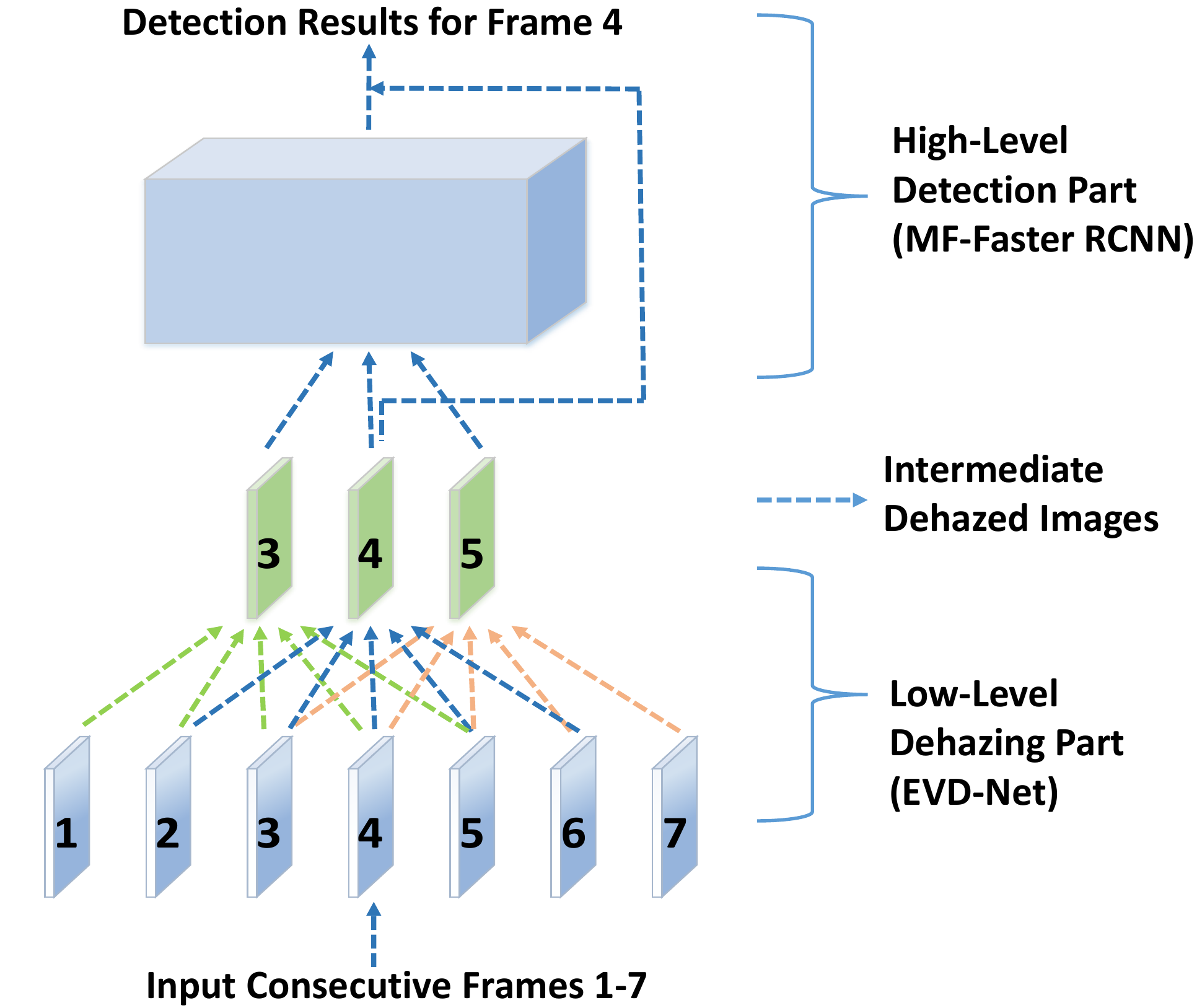}
	\caption{EVDD-Net for united video dehazing and detection, with a tree-like deep architecture. Note that the entire pipeline will be jointly optimized for training.}
	\vspace{-0.5em}
	\label{fig:tree_like}
\end{figure}

Recent years have witnessed a growing interest in modeling video using CNNs, for a wide range of tasks such as super-resolution (SR) \cite{kappeler2016video}, deblurring \cite{su2016deep}, classification \cite{karpathy2014large,shen2016fast}, and style transfer \cite{chen2017coherent}. \cite{kappeler2016video} investigates a variety of structure configurations for video SR. Similar attempts are made by \cite{karpathy2014large,shen2016fast}, both digging into different connectivity options for video classification. \cite{ICCV17b} proposes a more flexible formulation by placing a spatial alignment network between frames.\cite{su2016deep} introduces a CNN trained end-to-end to learn accumulating information across frames for video deblurring. For video style transfer, \cite{chen2017coherent} incorporates both short-term and long-term coherences and also indicates the superiority of multi-frame methods over single-frame ones.

\section{End-to-End Video Dehazing Network and Its Unity with Video Detection}

We choose the AOD-Net model \cite{ICCV17a,li2017all} for single image dehazing as the starting point to develop our deep video dehazing model, while recognizing that the proposed methodology can be applied to extending other deep image dehazing models to video, e.g., \cite{cai2016dehazenet,ren2016single}. Our main problem lies in the strategy of \textit{temporal fusion}. As a well-justified fact in video processing, jointly considering neighboring frames when predicting the current frame will benefit many image restoration and classification-type tasks \cite{liu2014bayesian,ma2015handling,kappeler2016video}. Specifically to the video dehazing case, both object depth (which decides the transmission matrix $T$) and the global atmospheric light $A$ should be hardly or slowly changed over a moderate number of consecutive frames, implying the great promise of exploiting multi-frame coherence for video dehazing.

\subsection{Fusion Strategy for Video Dehazing: Three Structure Options}

Enlightened by the analysis from \cite{kappeler2016video}, we investigate three different strategies to fuse consecutive frames. For simplicity, we show the architecture for five input frames as an example, namely the previous two $(t-2, t - 1)$, current $(t)$, and next two $(t + 1, t+2)$ frames, with the goal to predict the clean version for the current frame $t$. Clearly, any number of past and future frames can be accommodated. As compared in Figure \ref{fig:EVD-Net}, three different types of fusion structures are available for EVD-Net:
\begin{itemize}
\item \textbf{$I$-Level Fusion:} fusing at the input level. All five input frames  are concatenated along the first dimension before the first convolutional layer is applied. It corresponds to directly fusing image features at the pixel level, and then running single-image dehazing model on the fused image.
\item \textbf{$K$-Level Fusion:} fusing during the $K$-estimation. Specifically, we will term the following structure as \textit{$K$-level fusion, conv l} ($l$ = 1, 2, ... 5): each input frame will go through the first $l$ convolutional layers separately before concatenation at the output of the $l$-th layer, $l$ = 1, 2, ... 5. In other words, the multi-frame information is fused towards generating the key parameter $K$ (i.e., $t(x)$ and $A$) of the current frame, based on the underlying assumption that both object depths and global atmospheric light transmit smoothly across neighboring frames.
\item \textbf{$J$-Level Fusion:} fusing during the output level. It is equivalent to feed each frame to its separate $K$-estimation module, and the five $K$ outputs are concatenated right before the clean image generation module. It will not fuse until all frame-wise predictions have been made, and corresponds to fusing at the output level.
\end{itemize}
Training a video-based deep model is often more hassle. \cite{kappeler2016video} proves that a well-trained single-column deep model for images could provide a high-quality initialization for training a multi-column model for videos, by splitting all convolutional weights before the fusion step. We follow their strategy, training an AOD-Net first to initialize different EVD-Net architectures in EVD-Net.

\subsection{Unity Brings Power: Optimizing Dehazing and Detection as An End-to-End Pipeline in Video}

Beyond the video restoration purpose, dehazing, same as many other low-level restoration and enhancement techniques, is commonly employed as pre-processing, to improve the performance of high-level computer vision tasks in the presence of certain visual data degradations. A few pioneering works in single-image cases \cite{wang2016studying,ICCV17a,li2017all} have demonstrated that formulating the low-level and high-level tasks as one unified (deep) pipeline and optimizing it from end to end will convincingly boost the performance. Up to our best knowledge, the methodology has not been validated in video cases yet.

In outdoor surveillance or autonomous driving, object detection from video \cite{kang2016object,tripathi2016context,zhu2017flow} is widely desirable, whose performance is known to heavily suffer from the existence of haze. For example, autonomous vehicles rely on a light detection and ranging (LIDAR) sensor to model the surrounding world, and a video camera (and computer, mounted in the vehicle) records, analyzes and interprets objects visually to create 3D maps. However, haze can interfere with laser light from the LIDAR sensor and fail subsequent algorithms. 

In this paper, we investigate the brand-new joint optimization pipeline of video dehazing and video object detection. Beyond the dehazing part, the detection part has to take into account temporal coherence as well, to reduce the flickering detection results. With EVD-Net, we further design a video-adapted version of Faster R-CNN \cite{NIPS2015_5638} and verify its effectiveness, while again recognizing the possibility of plugging in other video detection models. For the first two convolutional layers in the classical single-image Faster R-CNN model, we split them into three parallel branches to input the previous, current, and next frames, respectively\footnote{The window size 3 here is by default, but could be adjusted.}. They are concatenated after the second convolutional layer, and go through the remaining layers to predict object bounding boxes for the current frame. We call it \textit{Multi-Frame Faster R-CNN} (\textbf{MF-Faster R-CNN}).

\begin{table}
	\caption{PSNR/SSIM Comparisons of Various Structures.}
	\small
	\begin{center}
		\begin{tabular}{ccc}
			\hline
			Methods&PSNR&SSIM\\
			\hline\hline
			$I$-level fusion, 3 frames&20.5551&0.8515\\
			$I$-level fusion, 5 frames&20.6095&0.8529 \\
			\hline	
			$K$-level fusion, conv1, 3 frames&20.6105&0.9076\\			
			$K$-level fusion, conv1, 5 frames&20.8240&\underline{0.9107}\\
			\hline
			$K$-level fusion, conv2, 3 frames&20.6998&0.9028\\
			\textbf{$K$-level fusion, conv2, 5 frames}&\underline{\textbf{20.9908}}&\textbf{0.9087}\\
			$K$-level fusion, conv2, 7 frames&20.7901&0.9049\\
			$K$-level fusion, conv2, 9 frames&20.7355&0.9042\\
			\hline
			$K$-level fusion, conv3, 3 frames&20.9187&0.9078 \\
			$K$-level fusion, conv3, 5 frames&20.7780&0.9051\\
			\hline
			$K$-level fusion, conv4, 3 frames&20.7468&0.9038\\
			$K$-level fusion, conv4, 5 frames&20.6756&0.9027\\
			\hline
			$K$-level fusion, conv5(K), 3 frames&20.6546&0.8999\\
			$K$-level fusion, conv5(K), 5 frames&20.7942&0.9046\\			
			\hline	
			$J$-level fusion, 3 frames&20.4116&0.8812\\
			$J$-level fusion, 5 frames&20.3675&0.8791\\
			\hline
		\end{tabular}
			\vspace{-0.5em}
	\end{center}
	\label{tab:structure_psnr}
\end{table}

\begin{table}
	\caption{PSNR/SSIM Comparisons of Various Approaches.}
	\small
	\begin{center}
		\begin{tabular}{ccc}
			\hline
			Methods&PSNR&SSIM\\
			\hline\hline
			ATM~\cite{sulami2014automatic}&11.4190&0.6534\\
			BCCR~\cite{meng2013efficient}&13.4206&0.7068\\
			NLD~\cite{berman2016non}&13.9059&0.6456\\
			FVR~\cite{tarel2009fast}&16.2945&0.7799\\		
			DCP~\cite{he2011single}&16.4499&0.8188\\		
			DehazeNet~\cite{cai2016dehazenet}&17.9332&0.7963\\	
			CAP~\cite{zhu2015fast}&20.4097&0.8848\\
			MSCNN~\cite{ren2016single}&20.4839&0.8690\\
			AOD-Net~\cite{ICCV17a}&20.6828&0.8549\\
			\hline
			STMRF~\cite{cai2016stmrf}&18.9956&0.8707\\
			\textbf{EVD-Net}&\textbf{20.9908}&\textbf{0.9087}\\
			\hline
		\end{tabular}
	\vspace{-1.5em}
	\end{center}
	\label{tab:PSNR}
\end{table}

Finally, uniting EVD-Net and MF-Faster R-CNN in one gives rise to EVDD-Net, which naturally displays an interesting locally-connected, tree-like structure and is subject to further (\underline{and crucial}) joint optimization. Figure \ref{fig:tree_like} plots an instance of EVDD-Net, with a \textit{low-level temporal window size} of 5 frames, and a \textit{high-level temporal window size} of 3 frames, leading to the \textit{overall temporal window size} of 7 frames. We first feed 7 consecutive frames (indexed at 1, 2, ..., 7) into the EVDD-Net part. By predicting on 5-frame groups with a stride size of 1, three dehazed results corresponding to the frames 3, 4, 5 will be generated. They are then fed into the MF-Faster R-CNN part to fuse the detection results of frame 4. Essentially, the tree-like structure comes from the two-step utilization of temporal coherence between neighboring frames, in both low level and high level. We are confident that such a tree-like structure will be of extensive reference values to more future deep pipelines that seek to jointly optimize low-level and high-level tasks.

\section{Experiment Results on Video Dehazing}
\subsubsection*{Datasets and Implementation}
We created a synthetic hazy video dataset based on (\ref{e1}), using 18 videos selected from the TUM RGB-D Dataset \cite{sturm12iros}, which captures varied visual scenes. The depth information is refined by the filling algorithm in \cite{silberman2012nyu}. We then split it into a training set, consisting of 5 videos with 100,000 frames, and a non-overlapping testing set called \textit{TestSet V1}, consisting of the rest 13 relatively short video clips with a total of 403 frames.

\begin{figure*}
	\centering
	\subfigure[Inputs]{
	\includegraphics[width=6.8in,height=0.8in]{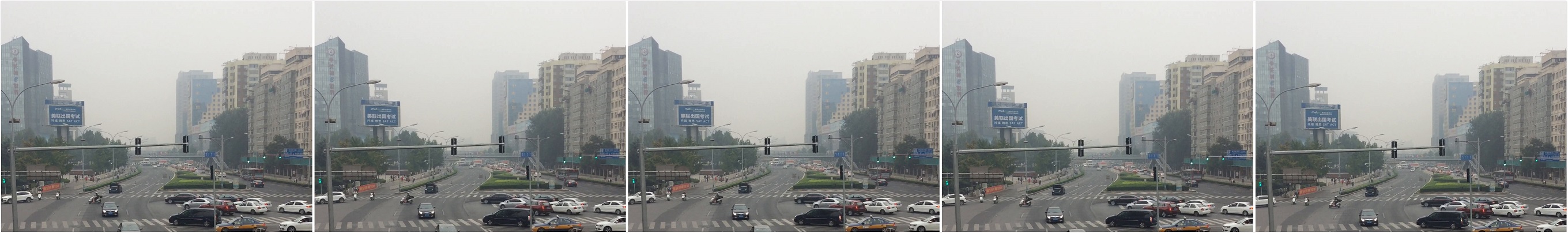}}
 	\subfigure[DCP]{
 	\includegraphics[width=6.8in,height=0.8in]{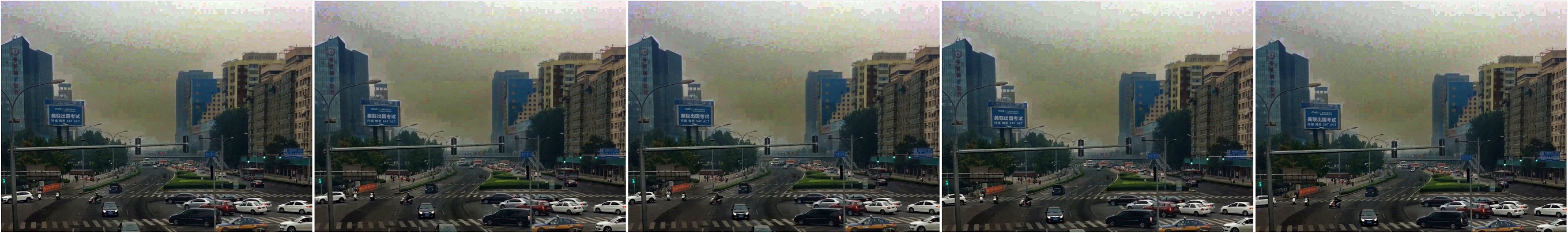}}
    \subfigure[NLD]{
 	\includegraphics[width=6.8in,height=0.8in]{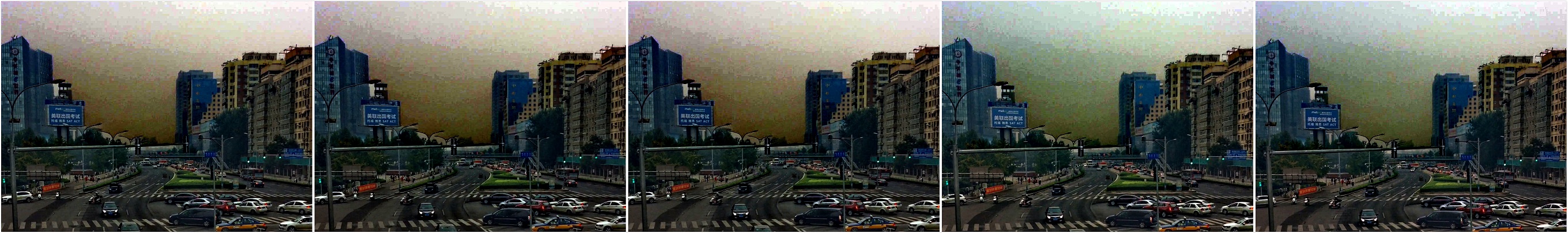}}
	\subfigure[CAP]{
	\includegraphics[width=6.8in,height=0.8in]{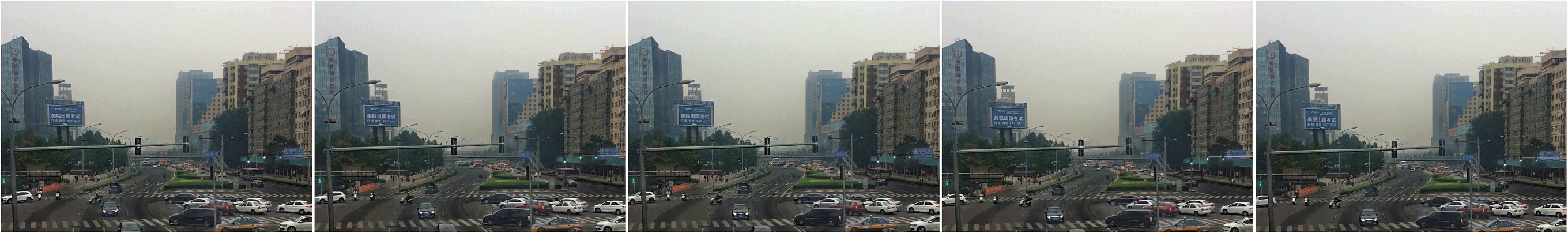}}
    \subfigure[MSCNN]{
	\includegraphics[width=6.8in,height=0.8in]{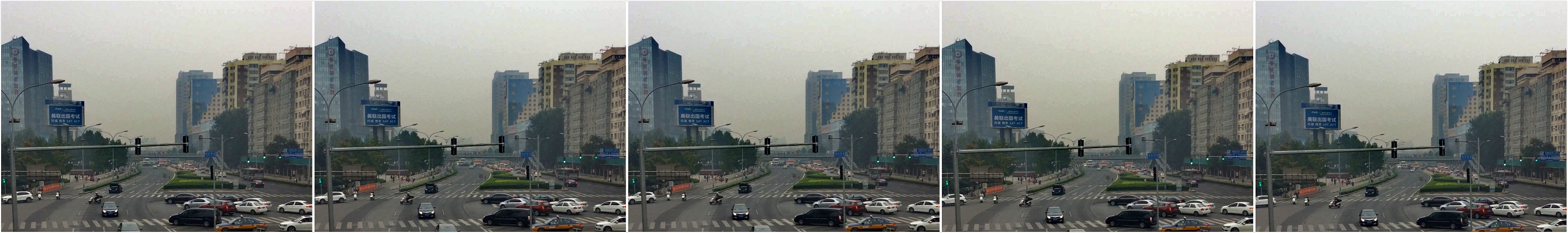}}
    \subfigure[DehazeNet]{
	\includegraphics[width=6.8in,height=0.8in]{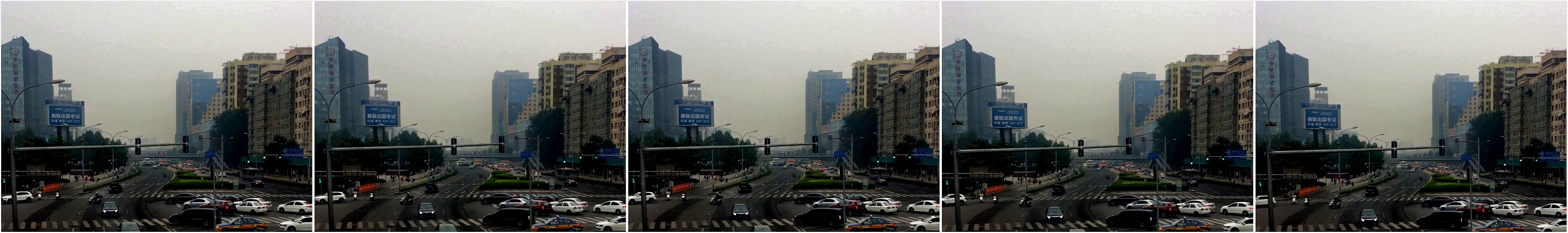}}
    \subfigure[AOD-Net]{
	\includegraphics[width=6.8in,height=0.8in]{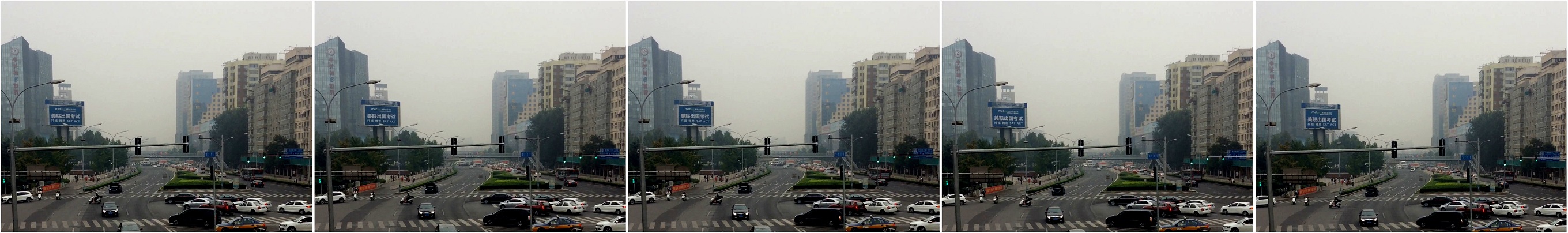}}
    \subfigure[STMRF]{
	\includegraphics[width=6.8in,height=0.8in]{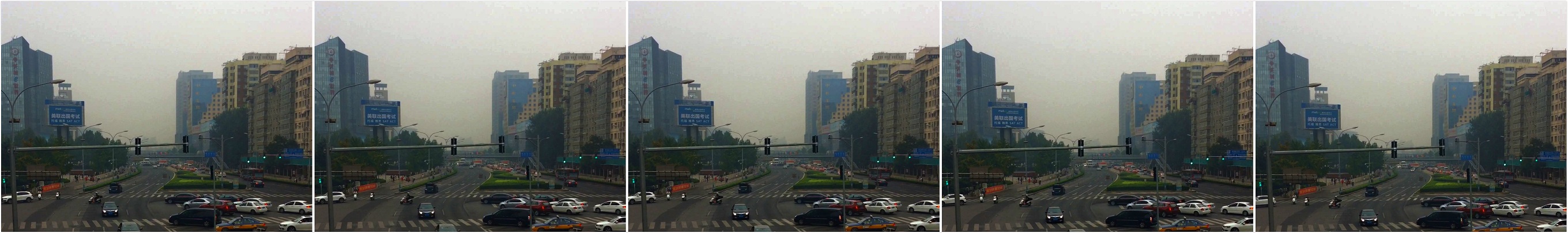}}
    \subfigure[EVD-Net]{
	\includegraphics[width=6.8in,height=0.8in]{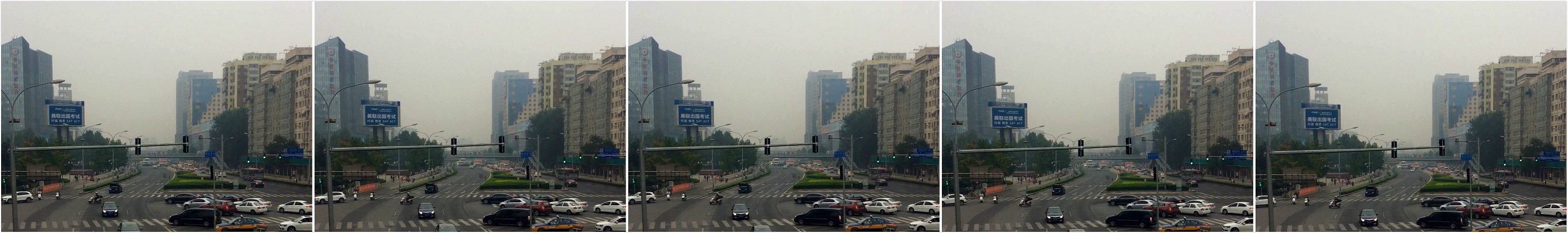}}
	\hspace{0.2in}
	\caption{Challenging natural consecutive frames results compared with the state-of-art methods.}
	\label{fig:vq}
\end{figure*}

\begin{figure*}
	\centering
	\subfigure[Inputs]{
		\includegraphics[width=1.1in,height=3in]{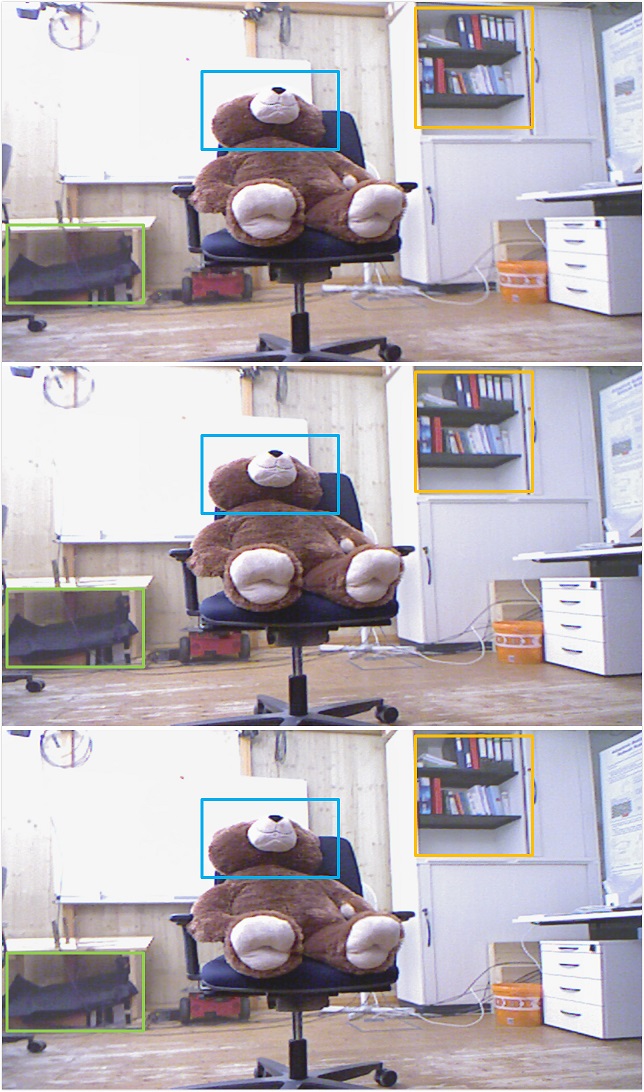}}
	\subfigure[MSCNN]{
		\includegraphics[width=1.1in,height=3in]{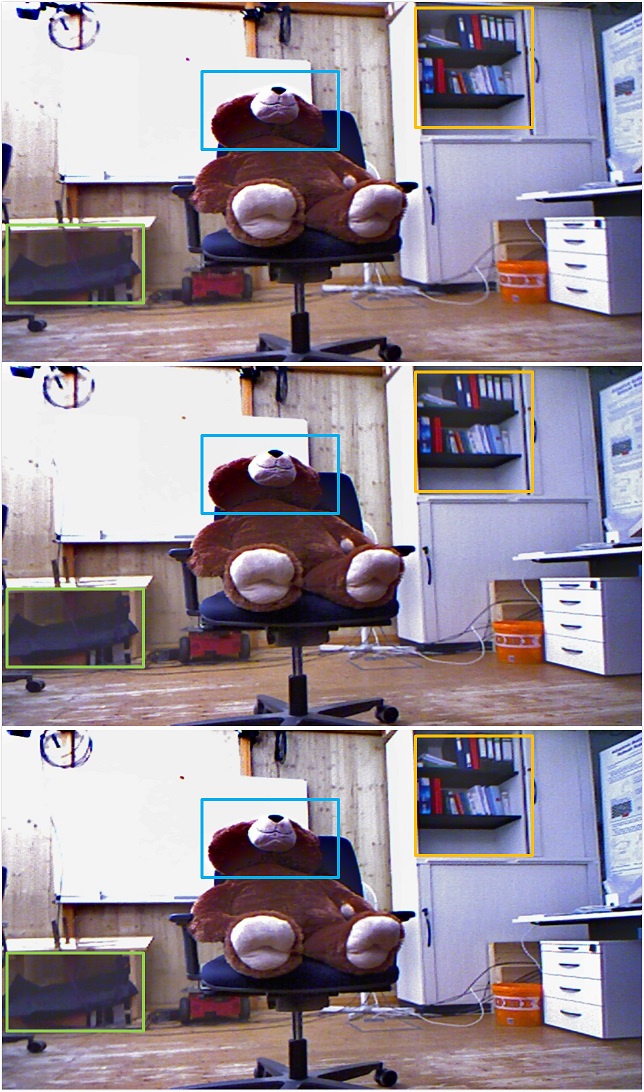}}
	\subfigure[DehazeNet]{
		\includegraphics[width=1.1in,height=3in]{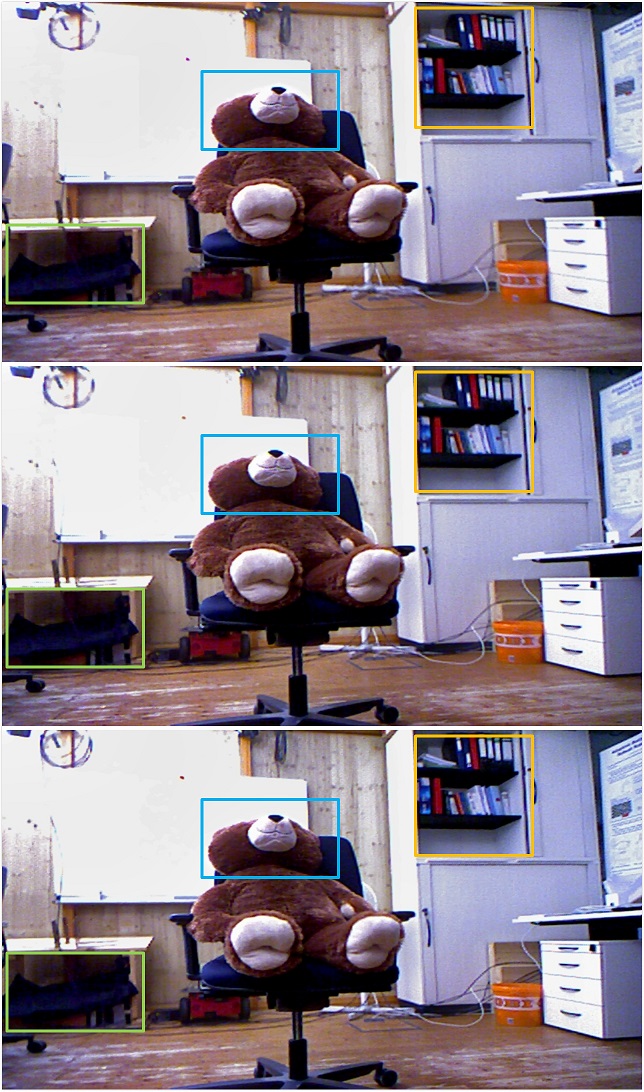}}
	\subfigure[STMRF]{
		\includegraphics[width=1.1in,height=3in]{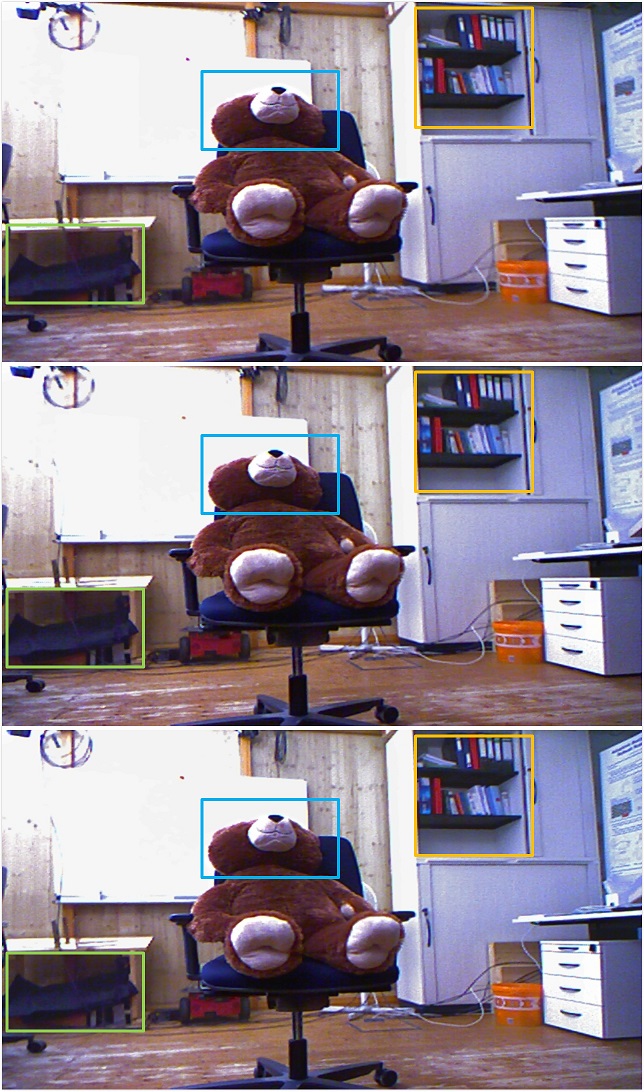}}
	\subfigure[EVD-Net]{
		\includegraphics[width=1.1in,height=3in]{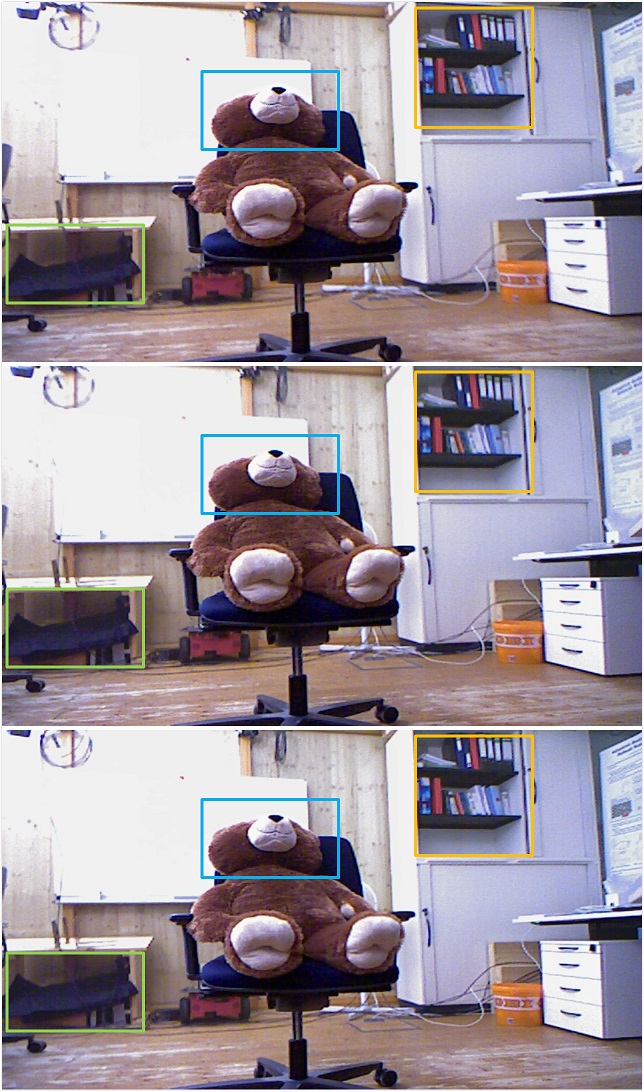}}
	\subfigure[Groundtruth]{
		\includegraphics[width=1.1in,height=3in]{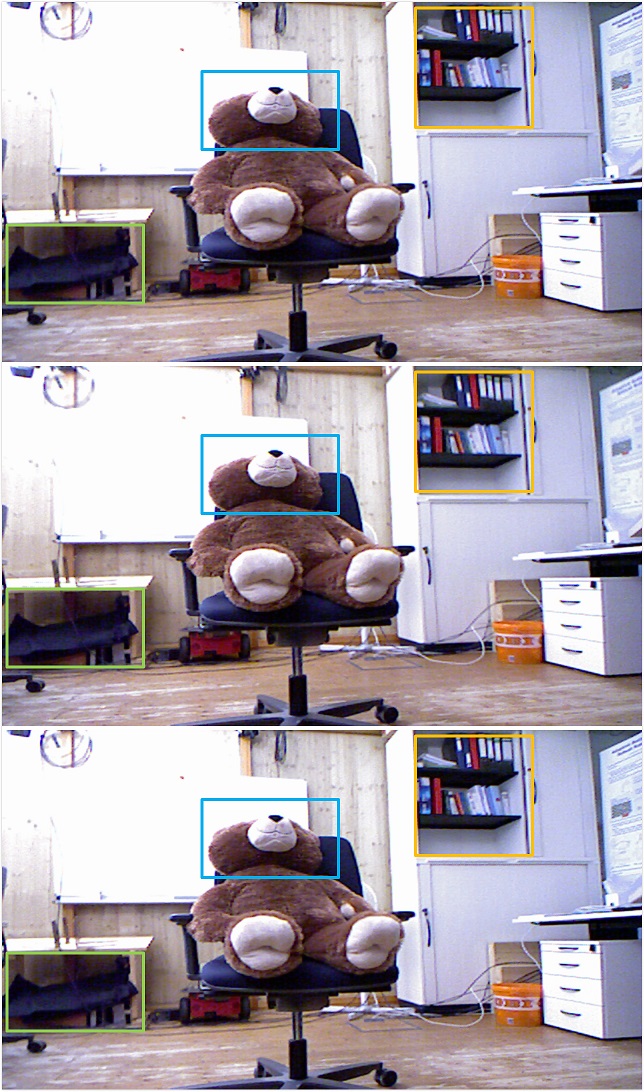}}
	\hspace{0.2in}
	\caption{Synthetic consecutive frames results compared with the state-of-art methods and Groundtruth frames.}
	\label{fig:vq1}
\end{figure*}

When training EVD-Net, the momentum and the decay parameters are set to 0.9 and 0.0001, respectively, with a batch size of 8. We adopt the Mean Square Error (MSE) loss, which has been shown in \cite{ICCV17a,li2017all} that it is well aligned with SSIM and visual quality. Thanks to the light-weight structure, EVD-Net takes only 8 epochs (100,000 iterations) to converge.

\begin{figure*}
	\centering
	\subfigure[Original Faster R-CNN]{
		\includegraphics[width=1.34in,height=3.8in]{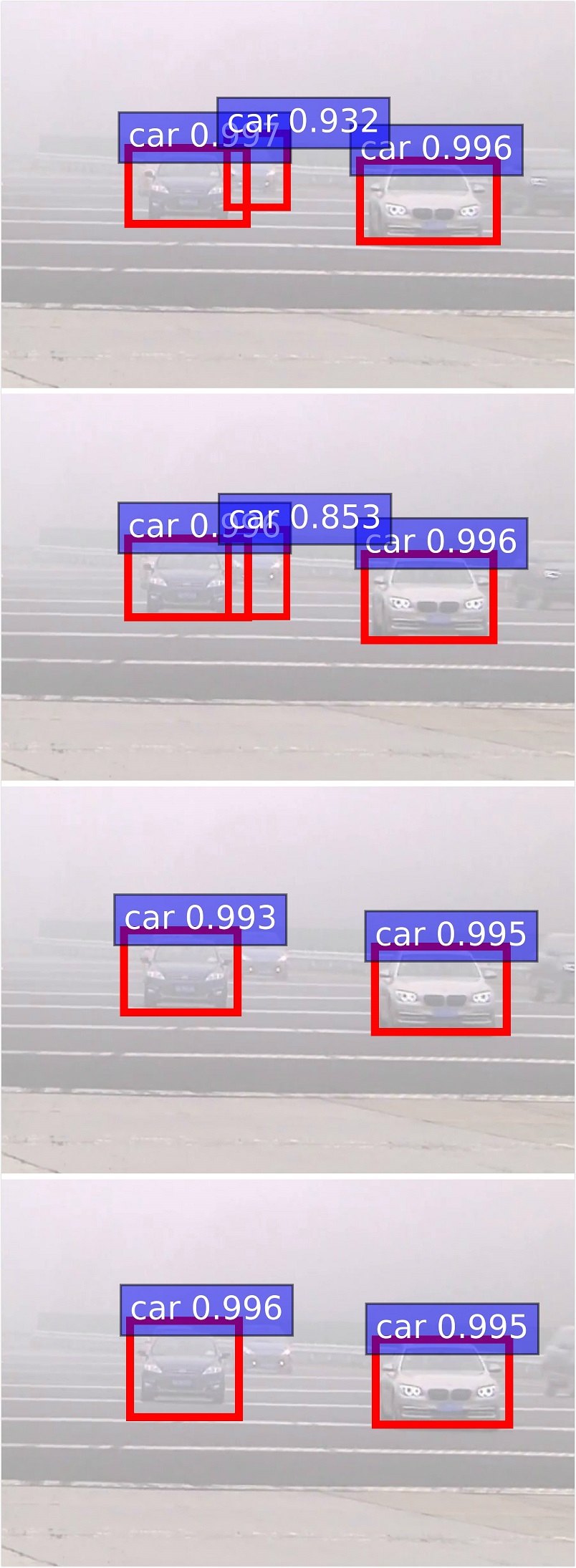}}
	\subfigure[\tiny{Retrained Faster R-CNN}]{
		\includegraphics[width=1.34in,height=3.8in]{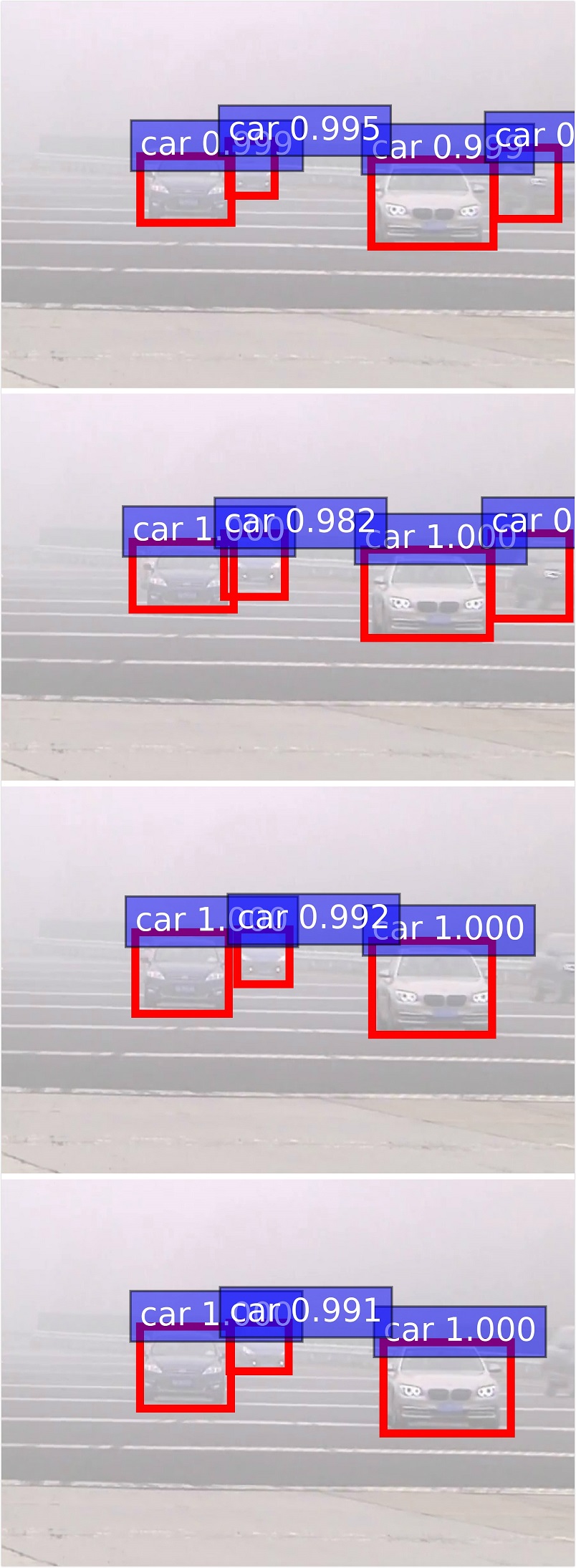}}
	\subfigure[EVD+Faster R-CNN]{
		\includegraphics[width=1.34in,height=3.8in]{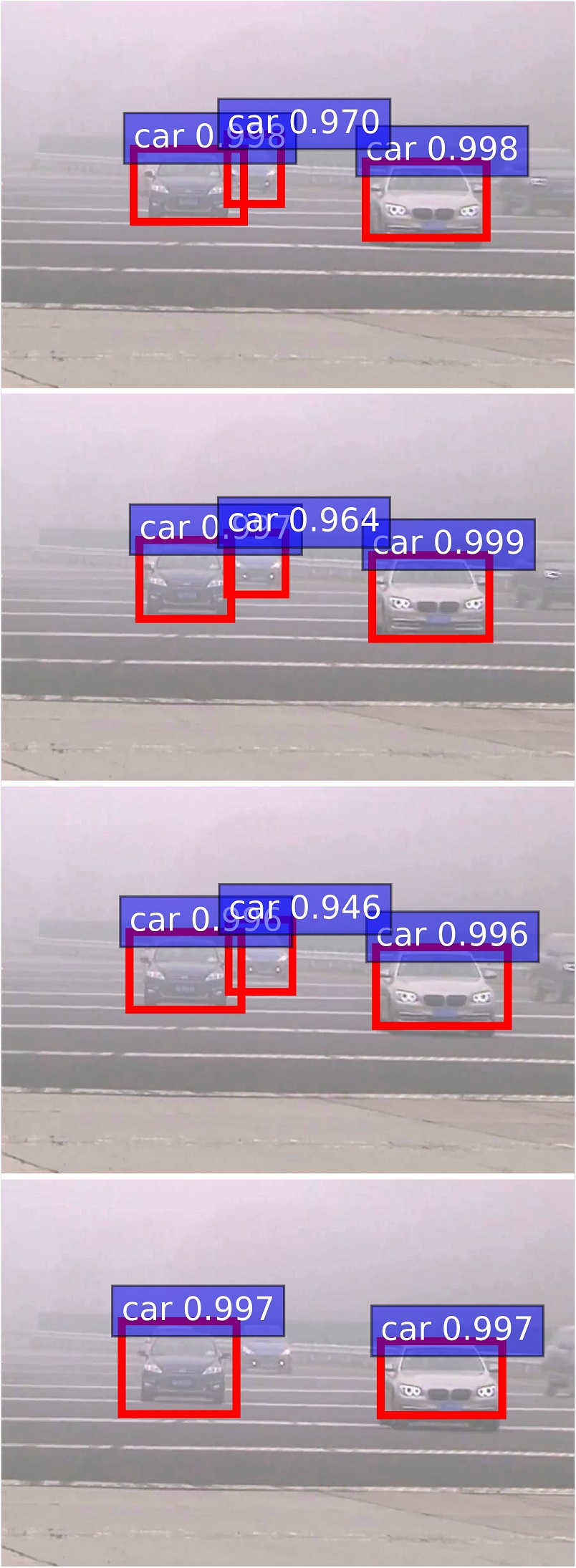}}
	\subfigure[JAOD-Faster R-CNN]{
		\includegraphics[width=1.34in,height=3.8in]{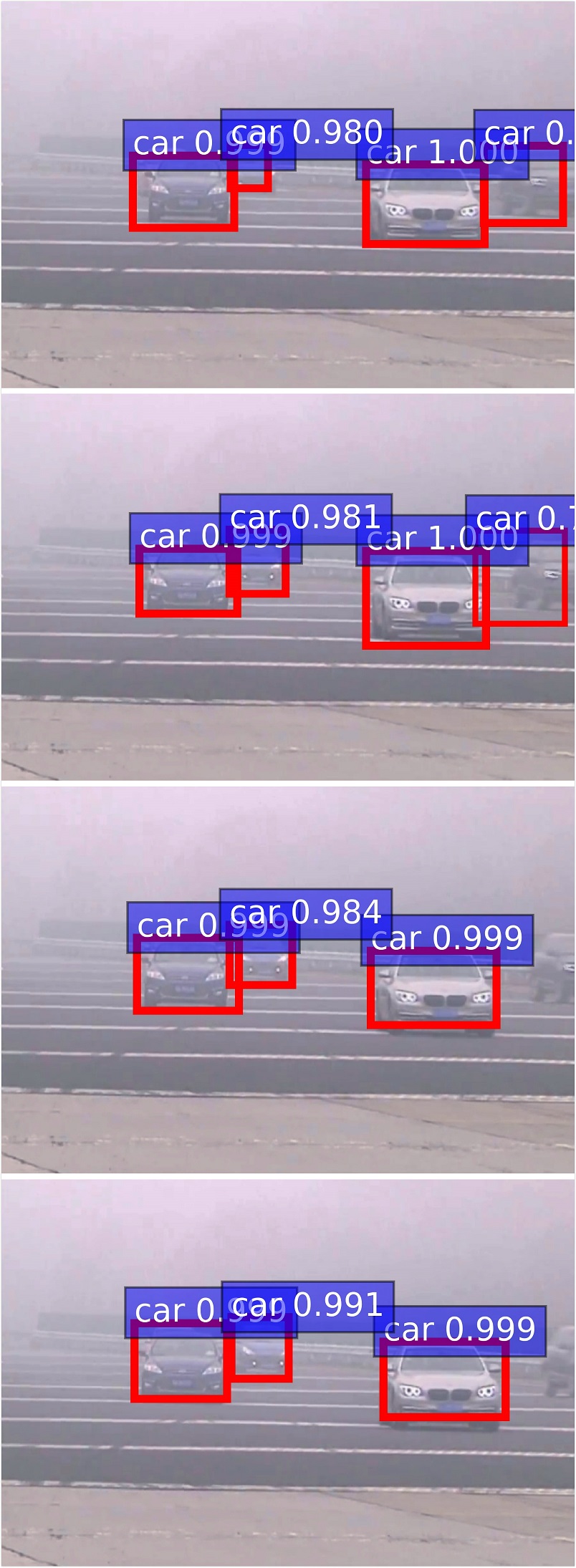}}
	\subfigure[EVDD-Net]{
		\includegraphics[width=1.34in,height=3.8in]{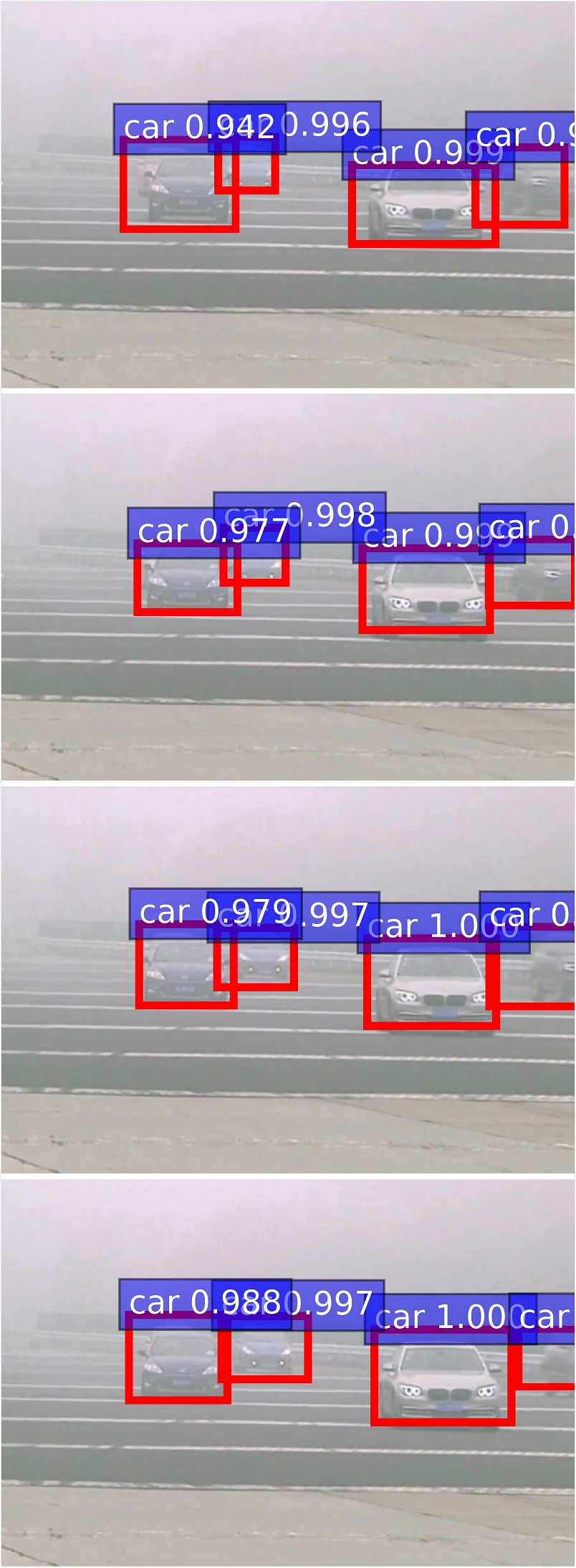}}
	\caption{Comparisons of detection results on real-world hazy video sample frames. Note that for the third, fourth and fifth columns, the results are visualized on top of the (intermediate) dehazing results.}
	\label{fig:real_detection}
\end{figure*}

\begin{table*}
	\caption{Average Precision(AP) of each categories and Mean Average Precision (MAP) on TestSet V2.}
	\vspace{-1em}
	\begin{center}
		\begin{tabular}{c|c|c|c|c|c}
			\hline
			Metrics&Original Faster R-CNN &Re-trained Faster R-CNN &EVD+Faster R-CNN&JAOD-Faster R-CNN&\textbf{EVDD-Net}\\
			\hline
			Car AP&0.810&0.807&0.811&0.808&0.803\\
			Bicycle AP&0.531&0.703&0.603&0.707&0.802\\
			MAP&0.671&0.755&0.707&0.758&\textbf{0.802}\\
			\hline
		\end{tabular}
			\vspace{-0.7em}
	\end{center}
	\label{tab:MAP}
\end{table*}

\subsubsection*{Fusion Structure Comparison}
We first compare the performances of three fusion strategies in EVD-Net with different configuration parameters on TestSet V1. As shown in Table \ref{tab:structure_psnr}, the performance of $K$-level fusion is far superior to $I$-level fusion and $J$-level fusion in both PSNR and SSIM, albeit the number of network parameters in $J$-level fusion is much more than the other two. Moreover, among all configurations of $K$-level fusion, when using 3 input frames, \textit{$K$-level fusion, conv 3} performs the best. While using 5 input frames the performance further increases and reaches an overall peak at \textit{$K$-level fusion, conv 2}, and is chosen as the \underline{default configuration of EVD-Net}. When testing more frames such as 7 or 9, we observe the performance gets saturated and sometimes hurt, since the relevance of far-away frames to the current frame will decay fast.

\subsubsection*{Quantitative Comparison}

We compare EVD-Net on TestSet V1 with a variety of state-of-the-art single image dehazing methods, including:  Automatic Atmospheric Light Recovery (\textbf{ATM})~\cite{sulami2014automatic}, Boundary Constrained Context Regularization (\textbf{BCCR})~\cite{meng2013efficient}, Fast Visibility Restoration (\textbf{FVR})~\cite{tarel2009fast}, Non-local Image Dehazing (\textbf{NLD})~\cite{berman2016non,berman2017air}, Dark-Channel Prior (\textbf{DCP})~\cite{he2011single}, \textbf{MSCNN}~\cite{ren2016single}, \textbf{DehazeNet}~\cite{cai2016dehazenet}, Color Attenuation Prior (\textbf{CAP})~\cite{zhu2015fast}, and \textbf{AOD-Net}~\cite{ICCV17a}. We also compare with a recently proposed video dehazing approach: Real-time Dehazing Based on Spatio-temporal MRF (\textbf{STMRF})~\cite{cai2016stmrf}. Table \ref{tab:PSNR} demonstrates the very promising performance margin of EVD-Net over others, in terms of both PSNR and SSIM. Compared to the second best approach of AOD-Net, EVD-Net gains an advantage of over 0.3 dB in PSNR and 0.05 in SSIM, showing the benefits of temporal coherence. Compared to the video-based STMRF (which is not CNN-based), we notice a remarkable performance gap of 2 dB in PSNR and 0.04 in SSIM.

\subsubsection*{Qualitative Visual Quality Comparision}

Figure~\ref{fig:vq} shows the comparison results on five consecutive frames for a number of image and video dehazing approaches, over a real-world hazy video (with no clean ground-truth). The test video is taken from a city road when the PM 2.5 is 223, constituting a challenging heavy haze scenario. Without the aid of temporal coherence, single image dehazing approaches tend to produce temporal inconsistencies and jaggy artifacts. The DCP and NLD results are especially visually unpleasing. CAP and MSCNN, as well as STMRF, fail to fully remove haze, e.g., in some building areas (please amplify to view details), while DehazeNet tends to darken the global light. AOD-Net produces reasonably good results, but sometimes cannot ensure the temporal consistencies of illumination and color tones. EVD-Net gives rise to the most visually pleasing, detail-preserving and temporally consistent dehazed results among all.

Figure~\ref{fig:vq1} shows a comparison example on synthetic data, where three consecutive frames are selected from TestSet V1. By comparing to the ground-truth, it can be seen that EVD-Net again preserves both details and color tones best.

\section{Experiment Results on Video Detection}

\subsubsection*{Datasets and Implementation}

While training EVDD-Net,   
the lack of hazy video datasets with object detection labels has again driven us to create our own synthetic training set. We synthesize hazy videos with various haze levels for a subset of ILSVRC2015 VID dataset \cite{russakovsky2015imagenet} based on the atmospheric scattering model in (\ref{e1}) and estimated depth using the method in \cite{liu2016learning}. The EVDD-Net is trained using 4,499 frames from 48 hazy videos for a two-category object detection problem (car, bike), and tested on 1,634 frames from another 10 hazy videos (\textit{TestSet V2}). Several real-world hazy videos are also used for evaluation. 

The training of EVDD-Net evidently benefits from high-quality initialization: a trained EVD-Net, plus a MF-Faster RCNN model initialized by splitting the first two convolutional layers similar to the way in \cite{kappeler2016video}. While \cite{li2017all} found that directly end-to-end training of two parts could lead to sufficiently good results, we observe that the video-based pipeline involves much more parameters and are thus more difficult to train end to end. Besides the initialization, we also find a two-step training strategy for EVDD-Net: we first tune only the fully-connected layers in the high-level detection part of EVD-Net for 90,000 iterations, and then tune the entire concatenated pipeline for another 10,000 iterations.

\subsubsection*{Comparison Baselines}
EVDD-Net is compared against a few baselines: i) the \textit{original Faster R-CNN} \cite{NIPS2015_5638}, which is single image-based and trained on haze-free images; ii) \textit{Re-trained Faster R-CNN}, which is obtained by retraining the original Faster R-CNN on a hazy image dataset; iii) \textit{EVD + Faster R-CNN}, which is a simple concatenation of separately trained EVD-Net and original Faster R-CNN models; iv) \textit{JAOD-Faster R-CNN}, which is the state-of-the-art single-image joint dehazing and detection pipeline proposed in \cite{li2017all}. 

\subsubsection*{Results and Analysis}
Table \ref{tab:MAP} presents the Mean Average Precision (MAP) of all five approaches, which is our main evaluation criterion. We also display the category-wise average precision for references. Comparing the first two columns verify that the object detection algorithms trained on conventional visual data do not generalize well on hazy data. Directly placing EVD-Net in front of MF-Faster R-CNN fails to outperform Retrained Faster-RCNN, although it surpasses the original Faster-RCNN with a margin. We notice that it coincides with some earlier observations in other degradation contexts \cite{wang2016studying}, that a naive concatenation of low-level and high-level models often cannot sufficiently boost the high-level task performance, as the low-level model will simultaneously bring in recognizable details and artifacts. The performance of JAOD-Faster R-CNN is promising, and slightly outperforms Retrained Faster-RCNN. However, its results often show temporally flickering and inconsistent detections. EVDD-Net achieves a significantly boosted MAP over other baselines. EVDD-Net is another successful example of ``closing the loop'' of low-level and high-level tasks, based on the well-verified assumption that the degraded image, if correctly restored, will also have a good identifiability. 

Figure \ref{fig:real_detection} shows a group of consecutive frames and object detection results for each approach, from a real-world hazy video sequence. EVDD-Net is able to produce both the most accurate and temporally consistent detection results. In this specific scene, EVDD-Net is the only approach that can correctly detect all four cars throughout the four displayed frames, especially the rightmost car that is hardly recognizable even for human eyes. That is owing to the temporal regularizations in both low-level and high-level parts of EVDD-Net. More video results can be found in the YouTube\footnote{https://youtu.be/Lih7Q91ykUk}.

\section{Conclusion}

This paper proposes EVD-Net, the first CNN-based, fully end-to-end video dehazing model, and thoroughly investigates the fusion strategies. Furthermore, EVD-Net is concatenated and jointly trained with a video object detection model, to constitute an end-to-end pipeline called EVDD-Net, for detecting objects in hazy video. Both EVD-Net and EVDD-Net are extensively evaluated on synthetic and real-world datasets, to verifyt the dramatic superiority in both dehazing quality and detection accuracy. Our future work aims to strengthen the video detection part of EVDD-Net.

{
	\small
	\bibliographystyle{aaai}
	\bibliography{egbib}
}

\end{document}